\documentclass{article}

\usepackage{PRIMEarxiv}

\usepackage[utf8]{inputenc} 
\usepackage[T1]{fontenc}    
\usepackage{hyperref}       
\usepackage{url}            
\usepackage{booktabs}       
\usepackage{amsfonts}       
\usepackage{nicefrac}       
\usepackage{microtype}      
\usepackage{lipsum}
\usepackage{fancyhdr}       
\usepackage{graphicx}       
\graphicspath{{media/}}     

\title{Understanding Lexical Biases when Identifying Gang-related
Social Media Communications}

\author{
  Dhiraj Murthy \\
  The University of Texas at Austin \\
  \texttt{dhiraj.murthy@austin.utexas.edu} \\
     \And
  Constantine Caramanis \\
  The University of Texas at Austin \\
  \texttt{constantine@utexas.edu} \\
 \And
  Koustav Rudra \\
  Northwestern University \\
  \texttt{koustav.rudra@northwestern.edu} \\
}

\begin{document}
\maketitle

\begin{abstract}
Individuals involved in gang-related activity use mainstream social media including Facebook and Twitter to express taunts and threats as well as grief and memorializing. However, identifying the impact of gang-related activity in order to serve community member needs through social media sources has a unique set of challenges. This includes the difficulty of ethically identifying training data of individuals impacted by gang activity and the need to account for a non-standard language style commonly used in the tweets from these individuals. Our study provides evidence of methods where natural language processing tools can be helpful in efficiently identifying individuals who may be in need of community care resources such as counselors, conflict mediators, or academic/professional training programs. We demonstrate that our binary logistic classifier outperforms baseline standards in identifying individuals impacted by gang-related violence using a sample of gang-related tweets associated with Chicago. We ultimately found that the language of a tweet is  highly relevant and that uses of ``big data'' methods or machine learning models need to better understand how language impacts the model's performance and how it discriminates among populations.
\end{abstract}


\section{Introduction}
Chicago has a  history of both high homicide rates as well as a  concentration of violence to particular neighborhoods \cite{dm_1}. Gang-related violence has been attributed to spikes in homicides, such as the 2016 rise from the baseline of 500 to 750 \cite{dm_2}. Gang related aggression can lead to a greater likelihood of mental health challenges for those individuals with high exposure to violence \cite{GangMembership_216,BadMedicine_217}, dangerous or reduced accessibility to schools \cite{Podcast_26}, and a cyclical continuation of violence through revenge \cite{GriefOnTwitter_8}. Furthermore, a narrative of violence can obscure the complexity of loss, grief, frustration, aggression, and perseverance experienced by individuals involved in or surrounded by gang networks \cite{ralph2014renegade}. 

Our work concerns the efforts of on-the-ground organizations to identify people impacted by gang violence as potential recipients of care resources. 
Community workers are on the frontline and, as such, are sometimes privy to possible violent incidents, and even their planned locations and times, before the police \cite{GangOutreach_93}. They can have unique opportunities to talk gang members out of constructing and sharing social media posts that incite violence. Other groups such as churches have leaders that mediate between gang members in times of conflict \cite{Englewood_35} or hold events like dance-offs that enable healthier forms of competition \cite{ralph2014renegade}, and schools monitor social media posts of gang affiliated students to evaluate security needs at large events like homecomings or prom \cite{Podcast_26}. 

We focused on the problem of identifying individuals on Twitter who have been impacted by gang related violence who could be ideal recipients of the care resources of these organizations. We also evaluated how our method is impacted by the language of the person tweeting, which provides insight for potential care workers on biases of the model.

\subsection{Existing Approaches}

When identifying individuals sharing a common experience on social media, such as a mental health challenge or expression of emotion, most previous work has focused on the inclusion of descriptive terms through hashtags or membership of a specific online community \cite{MentalHealthSignals_84,PTSD_85}. In the goal of identifying individuals involved in gang activity, previous work \cite{GangMembersTwitter_45} uses a similar method of selecting social media accounts with hashtags commonly used by gang members and pictures of illicit substances. However, there is little discussion of how this method could prove dangerous for those identified, especially given the potential criminal implications of gang-affiliation.

Other work identified patterns of loss and aggression in the tweets of gang affiliated individuals and created a model for classifying the two emotions displayed in these tweets \cite{GriefOnTwitter_8}. The literature highlights the challenge of understanding these tweets, which are frequently written in African American English (AAE) as opposed to the more  common Standard American English (SAE) \cite{YouthTwitterHelp_81}. This is reinforced by findings that demonstrate that widely used language processing tools perform more poorly on texts written with AAE than SAE when identifying the language of a tweet \cite{OffTheShelfTools_14} and when trying to identify parts of speech or identify synonyms of slang \cite{AutomaticallyProcessing_9}.

Recent work in the study of fairness and bias have made advances in evaluating how the discrepancies seen in the performance of these tools in language understanding impact text classification goals. Certain methods of substituting words differ along racial, political, and gender alignments and stylistic differences of similar words have been shown to cause unequal sentiment scores \cite{Darling_200}. Furthermore, there exist methods for identifying bias in classification problems that contain only text and no other traditional bias indicators like race, gender, or age by generating a synthetic test set for directly evaluating model bias \cite{Measuring_221}, applying an Equality of Odds based method for balancing false positive and false negative rates \cite{Equality_222}, and deriving a ``pinned AUC'' score to better evaluate bias among classes.

\subsection{Our Contribution}

We cast this problem as a binary classification of identifying whether an individual's tweets demonstrate a need for care resources to mitigate the impact of gang-related violence. This enables on-the-ground resources like school counselors, non-profit organizations, or social workers to identify people in need using open, online sources. It is our opinion that a substantially higher level of confidence is necessary when identifying people as impacted by gang violence compared to other situations, because any association with gang activity based on publicly accessible sources could have implications for policing or victimization even if this is not the intended purpose. We (1) demonstrate a method of gathering training data from a group of individuals who have been impacted by gang violence while avoiding an assumption of direct participation. This maintains an ethical standard for which predictive power of the model cannot be used to imply criminality. We then (2) describe a binary logistic regression classifier that more effectively identifies individuals impacted by gang-related violence in need of care resources that beats baseline standards.  Furthermore, we (3) define a method to evaluate how the language of a tweet impacts the bias of our task. This allows greater understanding of how the model targets specific communities in lieu of common metrics for fairness evaluation (like race, income, or wealth) that are not included in Twitter data, leading to greater interpretability for application uses.

\section{Data Sources}
\label{sources}

In order to build a classifier for identifying individuals impacted by gang violence and in need of community care resources and subsequently evaluate its performance on different forms of English language, we first collected tweets from a set of individuals determined to be impacted by gang violence and a set of random tweets in Chicago. We also selected tweets pretagged with emotion or style of English to later test the bias of the model.

\subsection{Challenges of Assuming Gang Affiliation}

Work by others in this area \cite{GangMembersTwitter_45} searches for hashtags like \#BDK (``Black Disciple Killers'') or \#FreeDaGuys and profiles with images containing individuals holding guns, with stacks of money, or showing gang signs in order to identify people with gang affiliations. However, using rules-based methods to identify gang members inherently poses higher than ideal uncertainty levels for determining gang membership. 

For example, we  found a Twitter user's profile that had a picture with a caption with the word ``Bloods'', a name of a well-known gang. We also saw that the user had images in their profile that contained an individual with large stacks of money close to them and that they had tweets that the word ``gang'' specifically. However, while this manual inspection revealed several references that suggested possible gang-affiliation, further research demonstrated the doubt in these assumptions. An open source list\footnote{https://chicagoganghistory.com/notorious-street-gangs/} of gangs in the user's expected location (hinted by geographic evidence in photographs) did not include the Bloods. The images of the person in their profile with a large stack of money contained captions with phrases conveying a longing to see the person again. This implies that  pictures may actually have been of someone different than the user or that the user might not have been present when the pictures were taken. We felt that due to the public nature of these posts, the method described above did not provide sufficient confidence in determining whether someone was gang affiliated. This highlights the need for more sophisticated and nuanced methods of classification.

\subsection{Ethics in Data Collection}

Because we examine tweets that share real experiences, and are partly composed of authors from marginalized populations,  we felt that it was important to develop an ethics statement to frame our work. This is especially important since our Institutional Review Board deems public Twitter data analytics exempt from approval and review. We chose to identify people in need of community care who did not have known criminal records. This means any predictions made could not be used as an indicator of criminality. We supplemented our technical knowledge with seminars about systemic causes of inequalities, relevant texts about living  and writing about an urban Chicago community, and conversations with leaders in broad academic disciplines including sociology and public policy who study the use of big data in the justice system. We have not shared data with anyone outside our research team. Finally, when presenting and writing about my work, any tweets from people not deceased are anonymized and altered so that they are unsearchable.

\subsection{Individuals Impacted by Gang Violence}

Given the challenges and implications with assuming gang-affiliation, we frame our task around identifying people who have been impacted by, rather than participated in, gang violence. This ``impact'' is defined by a personal relationship with the individual involved in the gang violence, leading to high susceptibility to loss and volatility overtime. People impacted by gang violence within this definition are often family members or close friends of individuals who have been physically harmed by or perpetrated gang related violence.

In order to identify people impacted by gang violence, we began with two well known gang-affiliated teenagers from urban Chicago who were killed in gang violence: Gakirah Barnes, whose tweets have been the source of a prominent study \cite{AutomaticallyProcessing_9} following her death in April of 2014, and Lamanta Reese, whose death in May 2017 was covered by Chicago news outlets \cite{LamantaSource_73}.

For Barnes, we used a set of 819 tweets retweeted by, mentioning, or authored by Barnes including the four months preceding her death and several weeks after.\footnote{https://academiccommons.columbia.edu/doi/10.7916/D84F1R07} These data include human labeling by a group of social workers that categorize the tweets as displaying either ``loss'', ``aggression'', or ``other'' \cite{AutomaticallyProcessing_9}. All of these tweets were designated as the positive lass of tweets, i.e. indicative of individuals impacted by gang violence. Due to limited access to long term Twitter stores for accessing specific users' tweets, we then scraped the maximum possible number of recent tweets retweeted by, mentioning, or authored by Reese on his public Twitter profile, resulting in 708 total tweets from more than five months before his death and several weeks after.

We then identified five people from each of the Twitter profiles of Barnes and Reese that most frequently appeared on their wall. We removed all accounts that had more than 2,000 followers unless they posted at some point directly mentioning Barnes or Reese in order to eliminate public figures who may not have personally known the two. Due to their relationship with someone who had died in gang related violence, these individuals fall within the definition of being ``impacted'' by gang violence and would have been ideal recipients of care resources. We then selected the top five people that most frequently mentioned Barnes or Reese and scraped their most recent publicly available tweets. This yielded an additional 7,000 tweets. We then selected only the tweets authored by the people who most frequently mentioned Barnes or Reese, excluding those that appeared on their walls but were not written by them (like mentions or retweets), for a total of approximately 2,800 additional tweets.

\subsection{Random Tweets from Chicago}

In order to identify a random population with an unknown need for community care resources, we extracted data from Twitter's `spritzer' stream API which delivered 1\% of all tweets that occurred over a one month period, from August 17, 2017 through September 17, 2017. We further isolated the tweets to a bounding box around the geographic area of Chicago, since both Barnes and Reese were from Chicago. We selected only tweets tagged by Twitter as written in English and removed retweets. This provided approximately 48,000 tweets.
\hfill{}\break{}
\hfill{}\break{}
\hfill{}\break{}

\begin{figure}
  \centering
  \includegraphics[width=90mm]{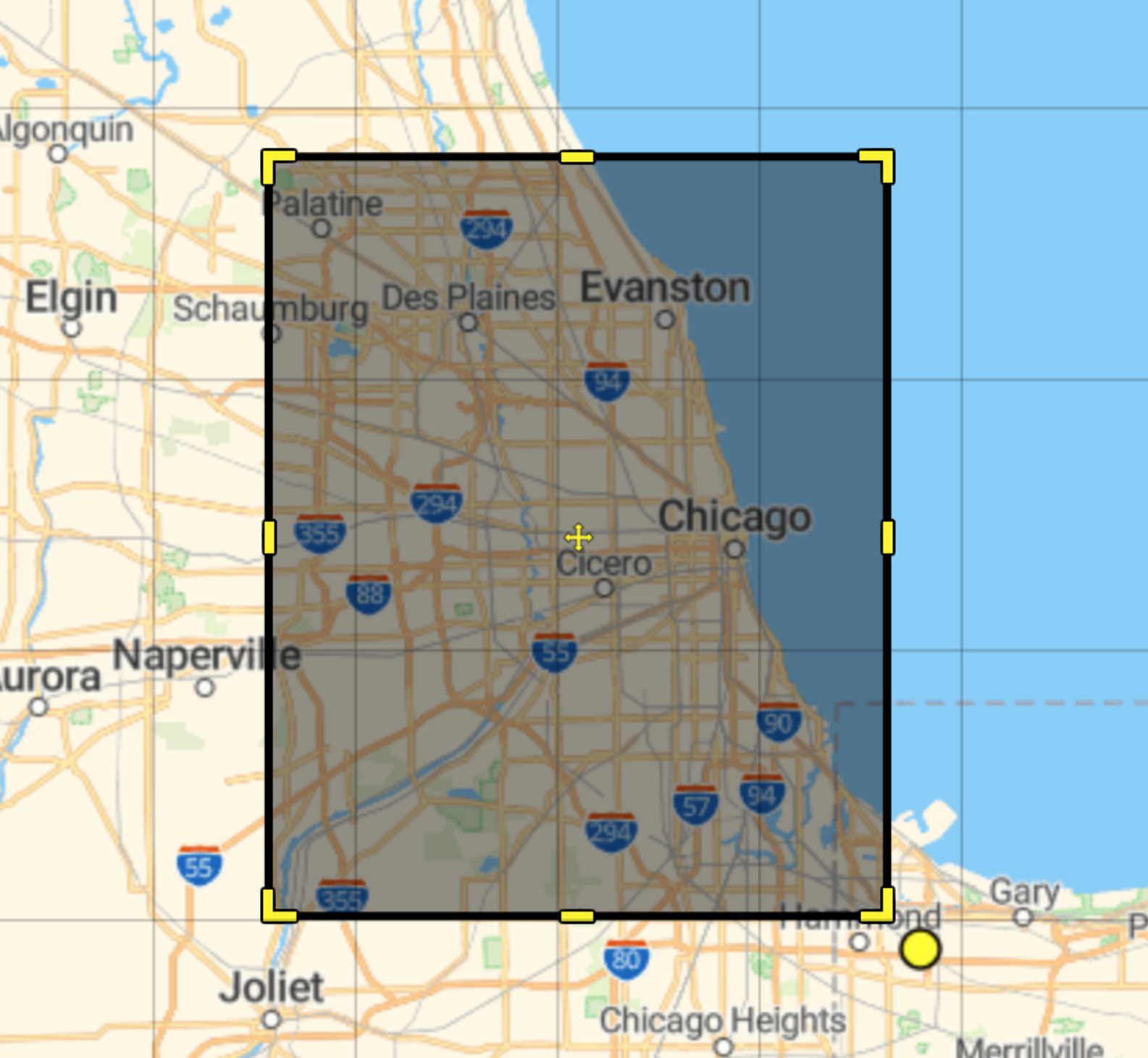}
  \caption{The bounding box defined for the random sample of tweets from Chicago.}
  \label{fig:bounding_box}
\end{figure}

\subsection{Tweets with Dialect Tags}

Established sociological work indicates long-term trends where African Americans in Chicago, due to ``housing discrimination, are differentially exposed to neighborhood conditions of extreme poverty'' \cite{dm_3}. And in these ghettoized Chicago neighborhoods, African American linguistic forms have played a long-standing, historical role in everyday life \cite{dm_4}. Previous work on tweets makes clear that ``urban gang involved youth [...] possess a unique style of communication that combines elements of Black vernacular English, language that is unique to their gang faction, and linguistic variation that is common on social media'' \cite{dm_5}.

To compare the performance of our classifier with different forms of English, we used an additional corpus of tweets from \cite{blodgett2016demographic} that have been machine labeled with a probability of being written by an African American, Hispanic, or White person, or none of the listed \footnote{http://slanglab.cs.umass.edu/TwitterAAE/}. The model was trained using text and geographic data. To build a dataset of tweets representative of African American English, we selected the tweets ``whose posterior probability of using AA [African American]-associated terms under the model was greater than 0.8''\cite{blodgett2016demographic}, yielding a total of 1.1 million tweets. 

\section{Experimentation}

We structured our model as a binary classifier for identifying whether an individual's tweets demonstrate a need for care resources to mitigate the impact of gang-related violence. The training data for those in need of care resources consisted of the dataset of tweets from individuals who had been impacted by gang violence (including Barnes, Reese, and the Twitter accounts that most frequently interacted with them, as explained in Section \ref{sources}) and the set of random tweets from Chicago served as the negative samples.

\subsection{Features and Model Structures}

We preprocessed all of the tweets using standard natural language processing methods including downcasing words andremoving stopwords and punctuation using the Natural Language Toolkit (NLTK)\footnote{https://www.nltk.org/}  library. We also used NLTK's Twitter-aware tokenizer module to reduce the length of words with redundant characters (such as ``waaaaayyyy too much'' to ``waaayyy too much''), remove account handles and tokenize the remaining tweet. Finally, we reduced all URLs containing the Twitter-shortened URL ``https://t.co/'' by removing trailing link identifiers. 

\begin{figure}
  \centering
  \includegraphics[width=60mm]{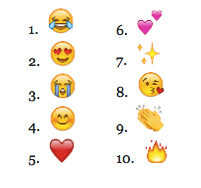}
  \caption{The 10 emojis most used on Twitter in 2015}
  \label{fig:top_emojis}
\end{figure}

We then used the preprocessed tweets to experiment with features. These included using text feature extraction libraries from scikit-learn\footnote{https://scikit-learn.org/stable/} to calculate term frequency-inverse document frequency at the word level, binary occurrence of words, and word bi-grams and utilizing NLTK's part-of-speech identifier to calculate the proportion of each tweet containing part-of-speech tags such as noun, adverb, adjective, and pronoun. We also counted the number of hashtags in each tweet, and the occurrences of each of the 10 emojis most used on Twitter in 2015 (as reported by Twitter Data \cite{PopularEmojis_79}), seen in Figure \ref{fig:top_emojis}. Furthermore, we created features based on the emotions associated with each word as found in the Emotion Lexicon\footnote{https://saifmohammad.com/WebPages/NRC-Emotion-Lexicon.htm}. We then calculated the proportion of each tweet that was made up of words displaying a given emotion for each anger, anticipation, disgust, fear, joy, sadness, surprise, or trust.

We then experimented with various models using scikit-learn's default Naive Bayes, binary logistic regression, and random forest, model structures and XGBoost's XGBClassifier model applied with combinations of the features.

\subsection{Predictive Performance}

Because there was a much smaller number of tweets from individuals in need of community care resources compared to the random sample, and we did not want obscure the performance of the model with regards to these tweets, we evaluated the performance of our models based on their measures for precision and recall. Precision is the measure of true positive tweets divided by the total number of tweets classified by the model as positive. A high precision would imply that an organization using the model could have greater confidence that a tweet classified by the model as being from an individual in need of community care resources is in fact demonstrative of a need for those resources. Recall is the measure of true positive tweets divided by the total number of true positive and false negative tweets. A higher recall would mean that an organization can have greater confidence that model has identified a large proportion of the tweets in the dataset that are actually indicative of individuals in need of community care resources.

Table \ref{tab:perf} shows the best performance for each of the models evaluated for several types of training and test splits, as well as two baselines for comparison, each of which are described further. The models in these tables used all of the features described in the previous section.

\subsubsection{Train and Test Splits}

We performed three different types of train/test splits to evaluate the precision and recall when classifying the tweets. In each of the methods, the random set of tweets from Chicago defining the negative class were randomly separated into test and training set. 

The \textbf{Tweet Level Method} trained models that contained a random, shuffled subset of the positive class and tested on the remaining tweets of the positive class, regardless of author. Scores are averaged over stratified (by predictive class), 5-fold classification testing using the scikit-learn StratifiedKFold module. This method demonstrates the model's success in identifying unseen tweets from known accounts whose other tweets were used to train the model. Its use case in real-life scenarios would be identifying tweets of unknown or disguised authorship (such as accounts whose screennames change in acts of memorializing or recognizing others, described in \cite{AutomaticallyProcessing_9}) in association with a small community known authors impacted by gang violence.

The \textbf{User Level Method} trained models on the tweets from all of the accounts found in the positive class except one, then tested on the tweets from the remaining user in the positive class. Scores are averaged over the 12 possible combinations in a ``leave-one-out'' fashion, in which each iteration leaves one of the accounts in the positive class for testing data. For each combination, a random sample of 20\% of the negative class was used as the test data for the negative class. This method conveys the efficacy of the model for identifying tweets from accounts not previously seen by the model but who share the same Twitter community connections to the original individual harmed in gang violence (in this case, Gakirah Barnes or Lamanta Reese) as the accounts used to train the model. It would be helpful in real-world situations for increasing the size of accounts in a known Twitter community of people impacted by gang violence.

The \textbf{Out-of-Network User Level Method} trained a model with the positive class made up of all of the tweets from the accounts that were identified through Gakirah Barnes' Twitter feed and tested with all of the tweets from accounts that were identified from Lamanta Reese's Twitter feed, or vice versa. The precision and recall scores are averaged over the two combinations of training and test splits. For each combination, a random sample of 20\% of the negative class is used as the test data for the negative class. This method demonstrates the model's ability to use a known community of individuals impacted by gang violence to identify a different group of people impacted by gang violence. It would be helpful in real-life situations for using Twitter to identify unknown, online communities impacted by gang violence. 

\subsubsection{Comparison to Baselines}

We defined two baselines for comparison of our model performance. These utilize rules-based methods similar to those discussed in Section \ref{sources}.

The \textbf{User Mention Baseline} identifies individuals impacted by gang violence by selecting all tweets that in their original, pre-processed form include a mention of either Gakirah Barnes' or Lamanta Reese's Twitter screennames as members of the positive class.

The \textbf{Aggression/Loss Annotation Baseline} utilizes an annotation guide \cite{Detecting_223} which maps specific words to labels of aggression or loss, as seen in Table \ref{tab:annotation}. For this baseline, any tweet that contains at least one of these words is selected as being indicative of an individual impacted by gang violence. 

\begin{table}[hbt!]
  \caption{Aggression/Loss Annotations}
  \centering
  \label{tab:annotation}
  \begin{tabular}{lp{50mm}}
    \toprule
    Label&Words\\
    \midrule
   	Aggression&angry, opps, opp, fu, fuck, bitch, smoke, pipe, glock, play, missin, bang, smack, slap, beat, blood, bust, bussin, heat, BDK, GDK, snitch, cappin, killa, kill, hitta, hittas, shooter, tf\\
    Loss&free, rip, longlive, LL, rest, up, restup, crying, cry, fly, flyhigh, fallin, bip,day, why, funeral, sleep, miss, king, hurt, gone, cant, believe, death, dead, died, lost, killed, grave, damn, soldier, soldiers, gang, bro, man, hitta, jail, blood, heaven, home\\
  \bottomrule
\end{tabular}
\end{table}
\hfill{}\break{}
\hfill{}\break{}
\hfill{}\break{}
\hfill{}\break{}
\hfill{}\break{}
\hfill{}\break{}

\begin{table}[hbt!]
  \caption{Model Performance}
  \centering
  \label{tab:perf}
  \begin{tabular}{cccl}
    \toprule
    Split Method&Model Type&Precision&Recall\\
    \midrule
    \textbf{Tweet Level} & Naive Bayes&90\%&39\%\\
     & Binary Log. Reg. &84\%&42\%\\
     & Random Forest &68\%&41\%\\
     & XGBoost &92\%&24\%\\
     & User Mention Baseline& 100\%& <1\%\\
     & Aggress/Loss Baseline& 8\%& 43\%\\
      \midrule
    \textbf{User Level} & Naive Bayes&54\%&29\%\\
     & Binary Log. Reg. &48\%&32\%\\
     & Random Forest &30\%&29\%\\
     & XGBoost &67\%&23\%\\
     & User Mention Baseline& 100\%& <1\%\\
     & Aggress/Loss Baseline& 3\%& 40\%\\
     \midrule
    \textbf{Out-of-Network} & Naive Bayes&95\%&14\%\\
     \textbf{User Level} & Binary Log. Reg. &93\%&20\%\\
     & Random Forest &81\%&20\%\\
     & XGBoost &97\%&17\%\\
     & User Mention Baseline& 100\%& <1\%\\
     & Aggress/Loss Baseline& 17\%& 42\%\\
  \bottomrule
\end{tabular}
\end{table}

\subsubsection{Discussion of Performance} Our experimentation demonstrated variations in performance among models and split levels.

\textbf{Model Performance.} As seen in Table \ref{tab:perf}, for each of the split methods, no single model using all features outperformed the others (and the two baselines) in both precision and recall. While the User Mention and Aggression/Loss baselines outperformed the models in only one of the two metrics (precision and recall, respectively) they dismally under-performed in the alternative metrics (recall and precision, respectively). This disparity could prove a significant challenge in real-life applications of attempting to allocate resources for those impacted by gang violence and highlights the overall improvement of each of the other models over the baselines. 

Specifically, the XGBoost model outperformed the other models in precision for all split levels while the Binary Logistic Regression classifier contained the highest (or equal) recall of the other models. Furthermore, the difference in performance between the XGBoost and Binary Logisitic Regression classifier models is much closer when XGBoost outperforms and much greater vice versa. Thus, while no model directly outperformed the others in both metrics, we claim that Binary Logistic Regression was the strongest performing model.

\textbf{Split Method Performance.} Across the three different splits, we observed that the Tweet Level method to have the greatest recall, meaning that it identifies more of the tweets from individuals impacted by gang violence. This is unsurprising, as the models are trained using tweets from the same authors that are featured in the test set. The greater precision of models using the Out-of-Network User Level method compared with  the models split using the User Level method may be explained in part by the Out-of-Network models comparably lower recall: the Out-of-Network models identify a smaller proportion of tweets from individuals impacted by gang violence, but have greater accuracy in the tweets they designate as such.

\section{Bias Evaluation}

Because data of individuals impacted by gang violence showed a higher proportion of tweets that were written in African-American English, we implemented a method for understanding how bias could be present in our training date, model, and output. 

\subsection{Feature Importances}

\begin{figure}[hbt!]
\includegraphics[width=4.0in]{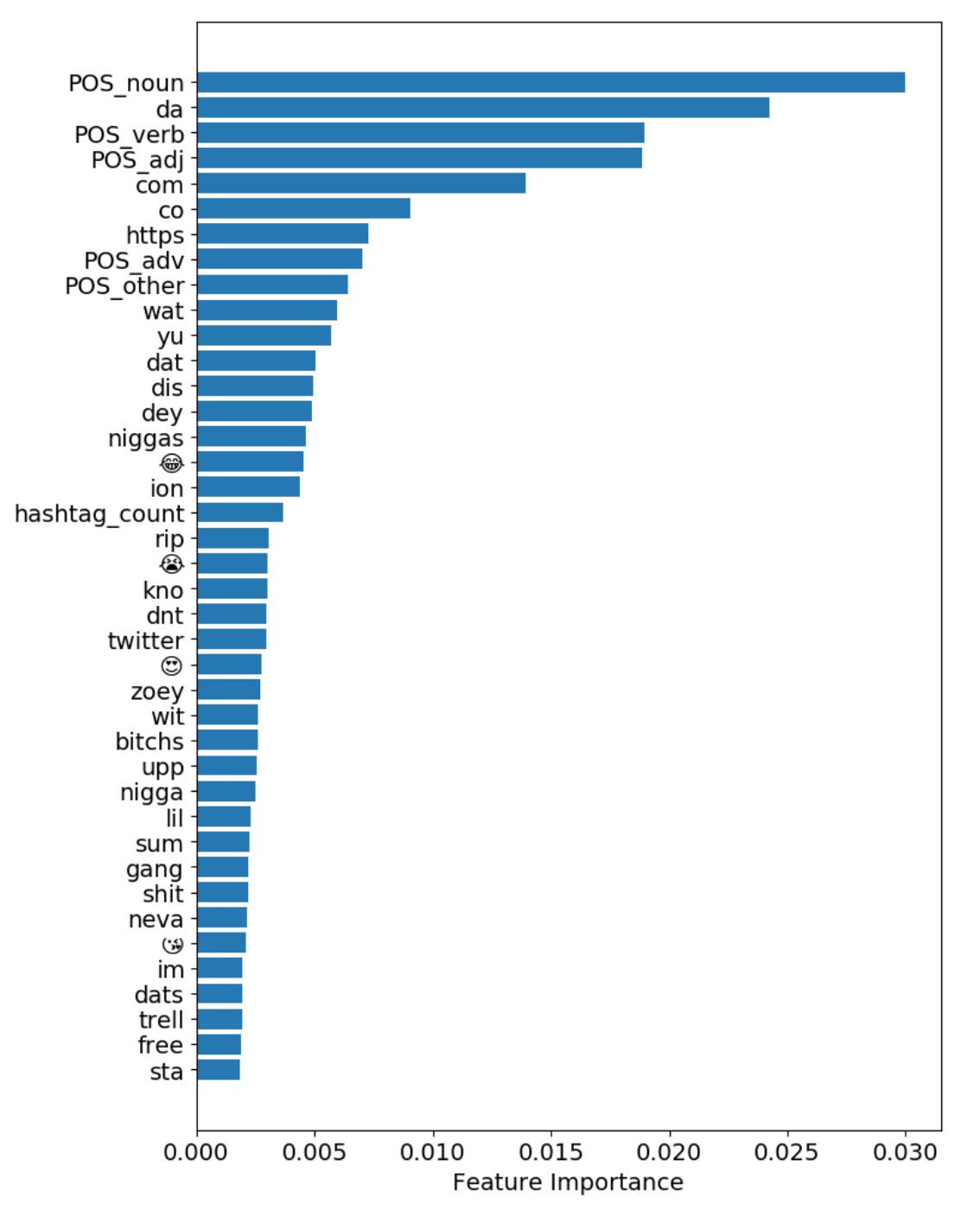}
\centering
\caption{Feature Importances for Random Forest Classifier with Tweet Level Split Method}
\label{fig:features}
\end{figure}

To better interpret how the models predicted individuals impacted by gang violence, we examined the feature importances of the model. We focused on the Random Forest classifier trained with the Tweet Level training/test split, which, it should be noted, was not the best performing model but has built in measures of feature importances. Although this model had a comparably high recall, it had a mediocre precision. This implies that if applying it, an organization would have lower confidence that the model was accurate in its designation of tweets authored by someone impacted by gang violence. To better understand the model, we first observed the most important one word and basic features of the model, as seen in Figure \ref{fig:features}.

We see that certain parts of speech features, emojis, and slang terms are very common among the top feature importances. The presence of these slang terms leads to uncertainty about whether the model prioritized features common to African American English.

\subsection{Forms of English and Features}

\begin{figure}
\includegraphics[width=4in]{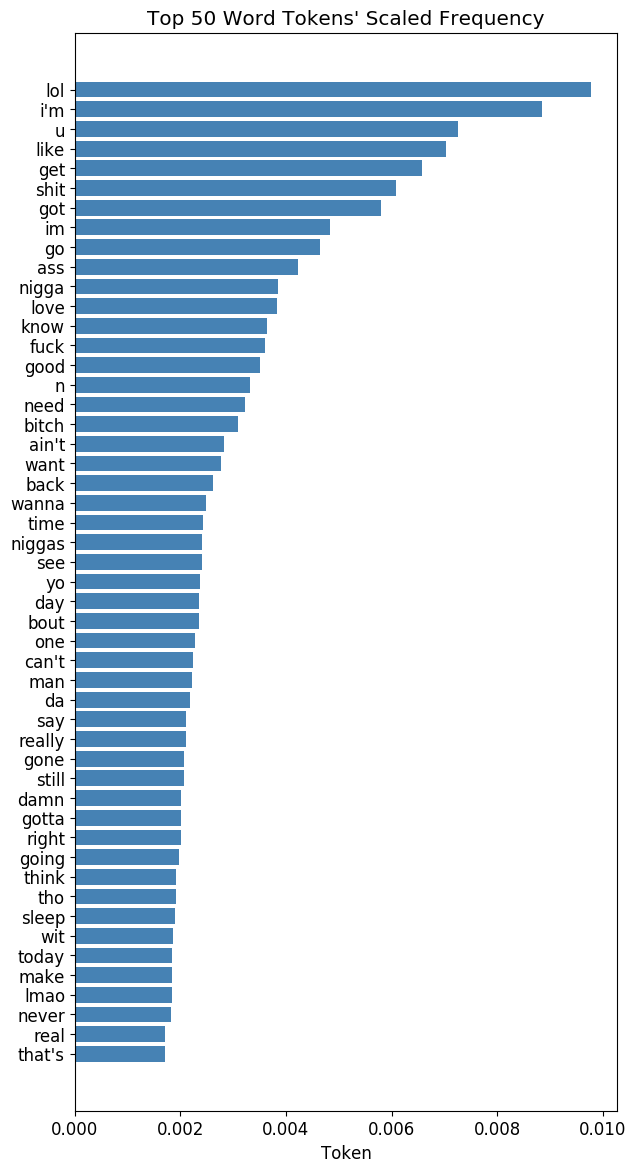}
\centering
\caption{Most Common Word Tokens From Tweets with High Probability of African American-Associated Terms}
\label{fig:word_freq}
\end{figure}

To understand the role of African American English in the model prediction, we evaluated how common each of these words were to the set of tweets that contained a high probability of containing African American-associated terms (described in Section \ref{sources}). We first preprocessed the tweets in the same way that we preprocessed the tweets used in identifying individuals impacted by gang. We then identified the 10,000 most common tokens in the dataset of African American-associated tweets. For each of the most common tokens, we counted the number of times it appeared in the corpus. For example, the word ``lol'' appeared more than 77,000 times amid the 1.1 million tweets, and the word ``like'' appeared more than 55,000 times. We then scaled each of the frequencies over the summed frequency counts for all 10,000 common tokens. Examples of the top 50 most common word tokens can be found in Figure \ref{fig:word_freq}.

Then, we bucketed the 7,500 most important single word features used in the aforementioned Random Forest classifier model consecutively into groups of 150 and summed their scaled frequency counts from the corpus of tweets with Afircan American-associated terms. 

\begin{figure}
\includegraphics[width=4in]{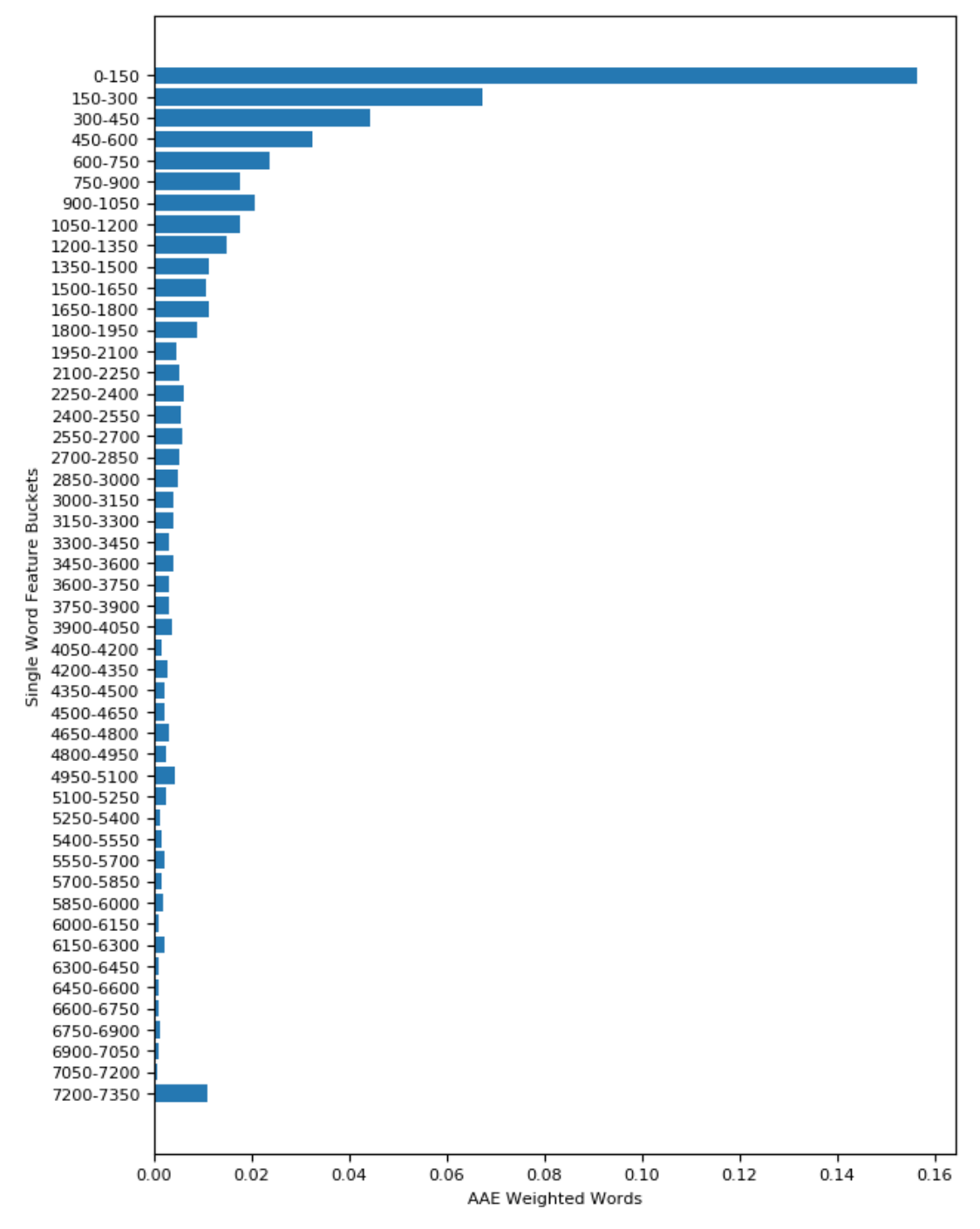}
\centering
\caption{Proportion of features appearing in AAE styled tweets}
\label{fig:featureword}
\end{figure}

Figure \ref{fig:featureword} is sorted from most to least important word features of the Random Classifier model, where each bucket of features contains the frequency of those 150 words within the most commonly used tokens in the African American-associated tweets. The figure demonstrates that more important features from the Random Classifier appear more frequently in the dataset of tweets that have a high probability of containing African American-associated terms. This finding is insightful for real world application because it quantifies the possibility of bias in the model and confirms the need to further evaluate the impact of lexical biases on the classifier's performance before implementation in real-life.

\section{Future Work}

The work discussed here can be used as a foundation for future development of predictive models or qualitative investigations into the opportunities for identifying individuals impacted by gang violence. 

We recommend the following areas for further investigation and research:

\begin{itemize}
  \item \textbf{Capture emoji meanings more effectively:} Emojis often express emotions and meanings not captured directly by text. Furthermore, certain communities attribute unique meanings to emojis. For example, recipients of higher rates of police misconduct could have a very different meaning for police emojis in their tweets than people who are tweeting about a local career day or an appreciation for public service. 
  \item \textbf{Evaluate emotions by tweet, rather than by word:} Due to restrictions on using more expensive APIs, we was only able to account for the emotions of individual words in tweets. Identifying an overall tweet emotion that accounts for the presence of multiple words in a  specific order could lead to a better understanding of tweet meaning.
  \item \textbf{Incorporate geographic data:} Tweet geo-locations could be insightful to certain challenges faced by the individual tweeting. For example, tweeting close to a site of a tragedy or crisis could provide better insight. Furthermore, vague tweets could be better understood if read in context with other tweets originating from a similar location.
  \item \textbf{Identify corroborating methods of lexical baselines:} The baseline for use of African American-associated terms in tweets could contain bias within itself. Using multiple data sources for determining styles of African American English and cross-referencing with other lexical styles could provide better insight to the true bias of the model.
\end{itemize}

\section{Conclusion}

Understanding the role of using social media to find individuals impacted by gang violence is a place where data science and sociological insights can be valuable, especially as most traditional  organizations have identified sources - even on social media -  manually, rather  than employing `big data' methods. These methods also seem to focus on people engaging in acts of gang violence, but not those in the periphery who are impacted by it and also need help. In this study, we demonstrate how a Binary Logistic Regression classifier using features including term frequency-inverse document frequency, binary occurrences of words, word bi-grams counts, proportion of parts-of-speech usage, hashtag counts, emoji counts, and proportion of emotion-related words outperforms existing, baseline methods of classification by placing specific emphasis on  understanding how models are influenced by differing forms of English. We conclude that individuals impacted by gang violence tend to tweet differently than others - different words, syntax, meaning. However, the style of tweeting could very well  be due to a user's demographic, educational, and community background. These variables clearly do not imply being impacted by gang violence. By identifying these lexical biases, we demonstrate a new, fairer method of identifying people who could receive care without an assumption  of gang involvement. Our work  also  underscores the need for future work to investigate how linguistic classifier models perform differently on different forms of language beyond  Standard American English and African American English.

\section*{Acknowledgments}
We acknowledge Melissa Hall’s efforts in leading the development and writing up of this study while a student at The University of Texas at Austin. The Plan II Honors program at the University of Texas at Austin provided financial support of Melissa Hall on this project. We thank Pranav Venkatesh [\texttt{pranav.venkatesh@utexas.edu}] for his help formatting plots, editing, and migrating our manuscript to arXiv.

\bibliographystyle{unsrt}  
\bibliography{references}

\end{document}